\documentclass[conference]{IEEEtran}
\IEEEoverridecommandlockouts
\usepackage{cite}
\usepackage{amsmath,amssymb,amsfonts}
\usepackage{algorithm}
\usepackage{algpseudocode}
\usepackage{subcaption}
\usepackage{graphicx}
\usepackage{textcomp}
\usepackage{xcolor}
\def\BibTeX{{\rm B\kern-.05em{\sc i\kern-.025em b}\kern-.08em
    T\kern-.1667em\lower.7ex\hbox{E}\kern-.125emX}}
\begin{document}

\title{Estimating Joint Probability Distribution With Low-Rank Tensor Decomposition, Radon Transforms and Dictionaries
\thanks{$^\dag$Pranava Singhal and Waqar Mirza are both first authors with equal contribution}
}

\author{\IEEEauthorblockN{Pranava Singhal$^\dag$}
\IEEEauthorblockA{\textit{Department of EE} \\
\textit{IIT Bombay}\\
Mumbai, India \\
\small{pranava.psinghal@gmail.com}}
\and
\IEEEauthorblockN{Waqar Mirza$^\dag$}
\IEEEauthorblockA{\textit{Department of EE} \\
\textit{IIT Bombay}\\
Mumbai, India \\
\small{wmirza608@gmail.com}}
\and
\IEEEauthorblockN{Ajit Rajwade}
\IEEEauthorblockA{\textit{Department of CSE} \\
\textit{IIT Bombay}\\
Mumbai, India \\
\small{ajitvr@cse.iitb.ac.in}}
\and
\IEEEauthorblockN{Karthik S. Gurumoorthy}
\IEEEauthorblockA{\textit{U.S. Omni Tech} \\
\textit{Walmart Global Tech}\\
Bangalore, India \\
\small{karthik.gurumoorthy@gmail.com}}
}

\newcommand{\alertbyKG}[1]{{\color{blue} Karthik:#1}}

\maketitle

\begin{abstract}
In this paper, we describe a method for estimating the joint probability density from data samples by assuming that the underlying distribution can be decomposed as a mixture of product densities with few mixture components. Prior works have used such a decomposition to estimate the joint density from lower-dimensional marginals, which can be estimated more reliably with the same number of samples. We combine two key ideas: dictionaries to represent 1-D densities, and random projections to estimate the joint distribution from 1-D marginals, explored separately in prior work. Our algorithm benefits from improved sample complexity over the previous dictionary-based approach by using 1-D marginals for reconstruction. We evaluate the performance of our method on estimating synthetic probability densities and compare it with the previous dictionary-based approach and Gaussian Mixture Models (GMMs). Our algorithm outperforms these other approaches in all the experimental settings.
\end{abstract}


\section{Introduction}
Joint probability density estimation is an important problem in several machine learning and statistical signal processing tasks \cite{intro1}\cite{intro2}. However, estimating the joint density from high dimensional data samples is challenging. Structure-free methods like histogramming and Kernel Density Estimation suffer from poor sample complexity. For instance, $N$ random variables (RVs), each taking $I$ distinct values, can take on $I^N$ distinct values of $N$-tuples. Reliable histogramming will need $\mathcal{S} \gg \Omega(I^N)$ samples since the probability of most $N$-tuples is usually quite small. Another approach is using graphical models assuming some conditional independence of RVs. However, such assumptions are problem specific and significantly restrict the kinds of densities that can be represented. 

Kargas \textit{et al.} \cite{kargas} proposed a framework to estimate the joint probability mass function of $N$ discrete RVs which represents the PMF (probability mass function) as a low-rank tensor using the Canonical Polyadic Decmposition (CPD). Following this work, many others \cite{truncfourier}\cite{sinc}\cite{jian}\cite{shaan} have tried to perform joint PMF and continuous PDF (probability density function) estimation. This entire line of work is connected by the idea that the CPD of the joint density shares factors with the CPD of its marginals. These low-dimensional marginals can be estimated reliably with much fewer samples and then used to obtain factors of the joint density.

In this paper we present a novel approach which combines the idea of dictionaries in \cite{shaan} with the use of Radon projections of 2D marginals to get 1D marginals in \cite{jian}. As a result, our work is the first one which can use the CPD framework to estimate continuous densities from 1D marginals. Dictionaries help us overcome restrictive assumptions like band-limitedness of the density used in \cite{sinc}. Moreover, sample complexity for obtaining 1D marginals of Radon projections is lower than that of 2D histogramming used in \cite{shaan}.

\section{Background}

\subsection{Canonical Polyadic Decomposition (CPD) of Tensors}
An $N$-dimensional tensor $\underline{\boldsymbol{X}} \in \mathbb{R}^{I_1\times I_2\times ... \times I_N}$ admits a `Canonical Polyadic Decomposition' as a sum of $F$ rank-1 tensors as follows:
$$\underline{\boldsymbol{X}} = \sum_{r = 1}^{F} \boldsymbol{\lambda}(r) \boldsymbol{A_1}(:, r) \circ \boldsymbol{A_2}(:, r) \circ ... \boldsymbol{A_N}(:, r),$$
where $F$ is the smallest number for which such a decomposition exists and is called the rank of the tensor. Here $\circ$ denotes the outer product of two vectors. For each $n \in [N] := \{1,2,...,N\}$, $\boldsymbol{A_n} \in \mathbb{R}_+^{I_n \times F}$ is called a mode factor and $\boldsymbol{\lambda} \in \mathbb{R}_+^F$. We also impose the constraint that $\forall n \in [N],\ \forall r \in [F]$, $\|\boldsymbol{A_n}(:,r)\|_1=1$.\\
If $\underline{\boldsymbol{X}}$ represents an N dimensional PMF, then we have the constraint that $\|\boldsymbol{\lambda}\|_1 = 1$. We can recover the PMF by estimating the mode factors and $\boldsymbol{\lambda}$ \cite{kargas}.

 We can extend the idea of CPD to a Naive-Bayes model for the PDF of an N-dimensional random vector $\boldsymbol{X} = (X_1, X_2, ..., X_N)$, where the random variables $\{ X_n\}_{n=1}^N$ are independent conditioned on a hidden variable $H$:
$$f_{\boldsymbol{X}}(x_1, x_2, ... , x_N) = \sum_{r = 1}^{F}f_{H}(r)\prod_{n = 1}^{N} f_{X_n|H}(x_n | H = r)$$
Here $f_X$ denotes the PDF of a random variable $X$.

\subsection{Joint PMF estimation from 2-way marginals}

In \cite{nmf}, Non-negative Matrix Factorisation (\textsc{Nmf}) techniques are employed to estimate the mode factors of the CPD of the PMF tensor from its 2D marginals. The 2D marginals $\boldsymbol{Z}_{j,k}$ are estimated using histogramming and we have the relation $\boldsymbol{Z}_{j,k} = \boldsymbol{A_j\Lambda A_k}^T$, where $\boldsymbol{\Lambda} = \textrm{diag}(\boldsymbol{\lambda})$. However \textsc{Nmf} techniques cannot be used if the tensor rank $F\gg \textrm{min}(I_j,I_k)$ since the mode factors will not be identifiable \cite{nmfidentify}. One can work around this by partitioning the set of indices of $N$ variables into two sets $\mathcal{S}_1 = \{l_1,l_2,...,l_M\}$ and $\mathcal{S}_2 = \{l_{M+1},l_{M+2},...,l_N\}$. Then a matrix $\boldsymbol{\Tilde{Z}}$ is constructed by concatenating blocks $\boldsymbol{Z}_{j,k}$ with $j \in \mathcal{S}_1$ (row-wise), $k \in \mathcal{S}_2$ (column-wise). We can then factorise $\boldsymbol{\Tilde{Z}}$ as $\boldsymbol{WH}^T$, where $\boldsymbol{W},\boldsymbol{H}$ can be obtained using the successive projection algorithm (\textsc{Spa}) \cite{nmfalgo} for \textsc{Nmf}. The mode factors are obtained from the relations $\boldsymbol{W}^T=[\boldsymbol{A}_{l_1}^T, \boldsymbol{A}_{l_2}^T,...,\boldsymbol{A}_{l_M}^T]$ and $\boldsymbol{H}^T=\boldsymbol{\Lambda}[\boldsymbol{A}_{l_{M+1}}^T,\boldsymbol{A}_{l_{M+2}}^T,...,\boldsymbol{A}_{l_N}^T]$.

\subsection{Dictionaries: Joint PDF estimation from 2-way marginals}
The work in \cite{shaan} extends the above technique to continuous densities. Each column of the continuous mode factors $f_{X_n|H}(x_n|H = r)$ is approximated by a convex combination of various 1D densities from a dictionary $\boldsymbol{\mathcal{D}_n}$. Thus we have the $r^{th}$ column
$$f_{X_n|H}(x_n|H = r) = \boldsymbol{\mathcal{A}_n}[:,r] = \boldsymbol{\mathcal{D}_n}\boldsymbol{B}_n[:,r]$$
where $1 \leq r \leq F$, $\boldsymbol{\mathcal{D}_n}$ is a dictionary of various continuous or discrete densities and $\boldsymbol{B}_n[:,r] \in \mathbb{R}_{+}^{L_n}$ is a non-negative weight vector which sums to one. $L_n$ is the number of density functions in the dictionary $\boldsymbol{\mathcal{D}_n}$. 

Then 1D marginals can be represented as $f_{X_n}(x_n) = \boldsymbol{\mathcal{D}_n}[x_n,:]\boldsymbol{B}_n\boldsymbol{\lambda}$, where $\boldsymbol{\mathcal{D}_n}[x_n,:]$ represent each of the density functions in $\boldsymbol{\mathcal{D}_n}$ evaluated at $x_n$ to yield a row vector. Since the 1D marginals are also convex combinations of the dictionary functions, the dictionary can be `guessed' by observing the  histograms of 1D marginals of data. 

The 2D marginals $\boldsymbol{\mathcal{Z}_{j,k}} = \boldsymbol{\mathcal{D}_j}\boldsymbol{B_j}\boldsymbol{\Lambda}\boldsymbol{B_k}^T\boldsymbol{\mathcal{D}_k}^T$, where calligraphic symbols represent continuous functions. The intervals $\{\Delta_n^{i_n}\}_{i_n=1}^{I_n}$ are used to bin the feature $X_n$. Discretised 2D marginals $\boldsymbol{{Z}_{j,k}}(i_j, i_k) = P(X_j \in \Delta_{j}^{i_j}, X_k \in \Delta_{k}^{i_k}) = \boldsymbol{{D}_j}\boldsymbol{B_j}\boldsymbol{\Lambda}\boldsymbol{B_k}^T\boldsymbol{D_k}^T$ are estimated by histogramming the data. Here the set of bins $\{\Delta_n^{i_n}\}_{i_n=1}^{I_n}$, also determines the discretisation used for columns of dictionary $\boldsymbol{D_n}$.

They estimate the mode factors by minimising the objective:
$$\sum_{j < k} \|\boldsymbol{\hat{Z}_{j,k}} - \boldsymbol{{D}_j}\boldsymbol{B_j}\boldsymbol{\Lambda}\boldsymbol{B_k}^T\boldsymbol{{D}_k}^T\|_{F}^{2}$$
$$\textrm{ s.t. } \, \forall j,r \, \|\boldsymbol{B_j}[:,r]\|_1 = \|\textrm{diag}(\boldsymbol{\Lambda})\|_1 = 1, \boldsymbol{B_j} \geq \boldsymbol{0}, \boldsymbol{\Lambda} \geq \boldsymbol{0}.$$
In the above equation, the $\geq$ sign acts element-wise. The use of dictionaries allows the representation of a larger variety of 1D marginals than is allowed by the bandlimitedness assumption in \cite{sinc}. The use of 2D marginal histograms lowers the sample complexity over using 3D marginals as in \cite{truncfourier}\cite{sinc}.

\subsection{Radon Transform: PMF estimation from 1-way marginals}
The work in \cite{jian} uses 1D marginals of the random projections of pairs of attributes values obtained from the high-D data. This is equivalent to computing Radon transforms of 2D marginal PMFs $\boldsymbol{Z_{j,k}}$ to get 1D marginals. The 1D marginals are estimated by histogramming data with lower sample complexity than 2D marginals.

For a pair of random variables $\boldsymbol{X}_{j,k} = (X_j, X_k) ,\ j<k$, and $M$ projection directions $\{\boldsymbol{\phi}_m \in \mathbb{R}^2 \, | \, m \in [M] \}$, we can estimate the 1D PMFs of the data projections $\langle \boldsymbol{\phi}_m,\boldsymbol{X}_{j,k} \rangle$ by histogramming into $I$ bins. Stacking the $M$ 1D PMFs row-wise yields matrix $\boldsymbol{Y_{j,k}} \in \mathbb{R}^{M\times I}$. They minimise the following objective:
$$\sum_{j<k} \|\boldsymbol{Y_{j,k}} - \mathcal{R}(\boldsymbol{A_j}\boldsymbol{\Lambda}\boldsymbol{A_k}^T)\|_{F}^{2}$$
$$\textrm{ s.t. }\, \forall n,r \, \|\boldsymbol{A_n}(:,r)\|_1 = \|\textrm{diag}(\boldsymbol{\Lambda})\|_1 = 1, \boldsymbol{A_n} \geq \boldsymbol{0}, \boldsymbol{\Lambda} \geq\boldsymbol{0},$$
where $\mathcal{R}$ represents the Radon operator. As discussed in \cite{nmf}, this objective can not be minimised directly since the mode factors will not be identifiable from 2D marginals. Thus, an auxiliary variable $\boldsymbol{Z_{j,k}}$ is introduced and the following unconstrained intermediate objective is optimised:
$$\sum_{j<k} \|\boldsymbol{Y_{j,k}} - \mathcal{R}(\boldsymbol{Z_{j,k}})\|_F^2 + \rho \|\boldsymbol{Z_{jk}}-\boldsymbol{A_j}\boldsymbol{\Lambda}\boldsymbol{A_k}^T\|_F^2.$$
The following procedure is followed: First $\boldsymbol{Z_{j,k}}$ is estimated by inverse Radon filtering $\boldsymbol{Y_{j,k}}$. Then $\boldsymbol{\Tilde{Z}}$ is assembled and \textsc{Spa} is used to estimate the mode factors as in \cite{nmf}.
Then the above intermediate objective is minimised with respect to $\boldsymbol{Z_{j,k}}$ followed by the \textsc{Spa} step to update mode factors, and this update is carried out multiple times. The mode factors obtained from the last step are used as the initialization to optimise the main objective.

\section{Problem Statement and Algorithm}
Given samples of an $N$-dimensional random vector $\boldsymbol{X} := (X_1, X_2, ..., X_N)$, we wish to estimate its probability density function $f_{\boldsymbol{X}}(x)$. We assume that each component $X_n$ may be continuous, discrete or mixed. We assume that the PDF admits a Canonical Polyadic Decomposition with some rank $F$, which is not known apriori. Similar to \cite{shaan}, we use dictionaries of densities $\boldsymbol{\mathcal{D}_n}$. We also utilise the Radon transform $\mathcal{R}(.)$ from \cite{jian} to project 2D marginals to obtain 1D marginals. We define the Radon projection operation as follows. For a given pair of features $(X_j,X_k) \,: j,k \in [N], j < k$ we generate projections in $M$ directions, with angles sampled uniformly from $[0,\pi)$. For a particular random projection vector $\phi_m$, we evaluate projections of all samples $\phi_m(x_j,x_k)^T$ and then histogram the projected values with a suitable bin size $b_m$. The obtained histogram density vectors for all directions are concatenated in a  tall vector of size $b:=b_1 + b_2 + ... + b_M$, denoted $\boldsymbol{y_{j,k}}$. 

For a chosen set of projection directions, the Radon operator $\mathcal{R}(.)$ is linear. We have the following relationship:
$$\boldsymbol{y_{j,k}} = \mathcal{R}(\boldsymbol{\mathcal{Z}_{j,k}}) = \mathcal{R}(\boldsymbol{\mathcal{D}_{j}}\boldsymbol{B_j}\boldsymbol{\Lambda}\boldsymbol{B_k^T}\boldsymbol{\mathcal{D}_k^T}).$$ Here, we introduce a new operator $\boldsymbol{\mathcal{R}_{j,k}}(.)$ defined as 
$$\boldsymbol{\mathcal{R}_{j,k}}(X) = \mathcal{R}(\boldsymbol{\mathcal{D}_{j}}X\boldsymbol{\mathcal{D}_k^T}) \, \forall X \in \mathbb{R}^{L_j \times L_k},$$
where $L_j$ denotes the number of columns in dictionary $\boldsymbol{\mathcal{D}_j}$. Thus we have 
$$\boldsymbol{y_{j,k}} = \boldsymbol{\mathcal{R}_{j,k}}(\boldsymbol{B_j}\boldsymbol{\Lambda}\boldsymbol{B_k^T}).$$

By linearity of $\mathcal{R}(.)$, even $\boldsymbol{\mathcal{R}_{j,k}}$ is linear and thus the operator can be replaced by a matrix multiplication as $\boldsymbol{\mathcal{R}_{j,k}}(X) = \boldsymbol{R_{j,k}}\cdot\text{vec}(X)$. Here $\text{vec}(X)$ denotes vertical concatenation of the columns of the input 2D matrix X to yield a tall vector. These matrices $\boldsymbol{R_{j,k}}$ can be pre-computed once the dictionaries have been chosen. The exact method of computing these matrices is discussed later in this paper. This is in contrast to \cite{jian} where the Radon transform is applied repeatedly rather than pre-computing the projection matrix once. While this does speed up the algorithm, a possible shortcoming of this approach is the large storage requirement of $N\choose 2$ pairs of projection matrices. The storage requirement can be considerably reduced by assuming identical dictionaries for all features and using the same projection directions.\\
In order to estimate $\{\boldsymbol{B_n} : n\in [N]\}$ and $\boldsymbol{\Lambda} = \text{diag}(\boldsymbol{\lambda})$ we optimise the following objective:\\
$$J_1({\boldsymbol{B_n}:n\in [N]}, \boldsymbol{\Lambda}) = \sum_{j<k} \|\boldsymbol{y_{j,k}} - \boldsymbol{\mathcal{R}_{j,k}}(\boldsymbol{B_j}\boldsymbol{\Lambda}\boldsymbol{B_k}^T)\|_{2}^{2}$$
$$\textrm{ s.t. }, \forall n,r \, \|\boldsymbol{B_n}(:,r)\|_1 = \|\textrm{diag}(\boldsymbol{\Lambda})\|_1 = 1, \boldsymbol{B_n} \geq \boldsymbol{0}, \boldsymbol{\Lambda} \geq \boldsymbol{0}.$$
To handle identifiability issues, we introduce an auxiliary variable $\boldsymbol{T_{j,k}}$ and an intermediate objective similar to that in \cite{jian}, as follows:
$$J_2(\boldsymbol{T_{j,k}}:j,k>j \in [N], {\boldsymbol{B_n}:n\in [N]}, \boldsymbol{\Lambda}) = $$
$$\sum_{j<k} \|\boldsymbol{y_{j,k}} - \boldsymbol{{R}_{j,k}}. \text{vec}(\boldsymbol{T_{j,k}})\|_2^2 + \rho \|\boldsymbol{T_{j,k}}-\boldsymbol{B_j}\boldsymbol{\Lambda}\boldsymbol{B_k}^T\|_F^2$$
$$\textrm{ s.t. }, \forall n,r \, \|\boldsymbol{B_n}(:,r)\|_1 = \|\textrm{diag}(\boldsymbol{\Lambda})\|_1 = 1, \|\text{vec}(\boldsymbol{T_{j,k}})\|_1 = 1$$
$$ \boldsymbol{B_n} \geq 0, \boldsymbol{\Lambda} \geq \boldsymbol{0}, \text{vec}(\boldsymbol{T_{j,k}}) \geq \boldsymbol{0}.$$
The algorithm proceeds as follows:\\
First $\boldsymbol{T_{j,k}}$ is initialised using $\boldsymbol{y_{j,k}}$ by minimising $J_2$ with $\rho = 0$ (simplex constrained least squares estimate).\\
Then $\boldsymbol{\Tilde{T}}$ is assembled and \textsc{Spa} is used to estimate the mode factors $\{ \boldsymbol{B_n} : n\in [N]\}$ and $\boldsymbol{\Lambda}$ as discussed in \cite{nmf}.\\
Next, the following alternating minimisation is carried out for several iterations: $J_2$ is minimised with respect to $\boldsymbol{T_{j,k}}$ keeping $\{ \boldsymbol{B_n} : n\in [N]\}$ and $\boldsymbol{\Lambda}$ fixed, followed by the \textsc{Spa} step to update $\{ \boldsymbol{B_n} : n\in [N]\}$ and $\boldsymbol{\Lambda}$ from the assembled $\boldsymbol{\Tilde{T}}$ matrix.\\
The mode factors obtained from the alternating minimisation are used as an initialisation in the minimisation of the main objective $J_1$, which is carried out using projected gradient descent. We name our algorithm \textsc{Rad} (\textbf{R}adons \textbf{A}nd \textbf{D}ictionaries). The pseudocode for \textsc{Rad} is given below.

\begin{algorithm}
	\caption{\textbf{RAD}: Estimating Mode Factors from 1D Random Projections Using Dictionaries}
	\begin{algorithmic}[*]
            \State Initialise dictionaries by hand based on the 1D marginal histograms
            \For {each pair $(j,k), j<k$}
                \State Generate $M$ random projection directions
                \State $\boldsymbol{y_{j,k}} \gets $ histograms of projected samples
                \State Compute $\boldsymbol{R_{j,k}}$ matrix
                \State Initialise $\boldsymbol{T_{j,k}}$ $\gets$ $\arg\min \|\boldsymbol{y_{j,k}} - \boldsymbol{{R}_{j,k}}. \text{vec}(\boldsymbol{T_{j,k}})\|_2$ under simplex constraint
            \EndFor
            \State converged $\gets$ False, $q \gets 1$
            \While {not converged}
                \State Assemble $\boldsymbol{\Tilde{T}}$ from $\{\boldsymbol{T_{j,k}}\}$
                \State $\boldsymbol{\{\boldsymbol{B_n}\}},\boldsymbol{\lambda}$ $\gets$ $\textsc{Spa}(\boldsymbol{\Tilde{T}})$
                \For {each pair $(j,k), j<k$}
                    \State $\boldsymbol{T_{j,k}}$ $\gets$ $\arg\min \|(\boldsymbol{{R}_{j,k}}^T\boldsymbol{y_{j,k}} + \text{vec}(\boldsymbol{B_j}\boldsymbol{\Lambda}\boldsymbol{B_k}^T) - (\boldsymbol{{R}_{j,k}}^T\boldsymbol{{R}_{j,k}} + \rho I). \text{vec}(\boldsymbol{T_{j,k}})\|_2$ under simplex constraint
                \EndFor
                \State $J_{q} \gets J_2(.)$, $q \gets q + 1$
                \If {$|J_{q} - J_{q-1}| < \epsilon$ OR $q == \text{max iterations}$}
                    \State converged $\gets$ True
                \EndIf
            \EndWhile
            \State converged $\gets$ False, $q \gets 1$
            \While {not converged}
            \Comment{choose $\eta$ using Armijo Rule}
                \For {$n \in [N]$}
                    \State $\boldsymbol{B_n} \gets \text{ProjectOntoSimplex}(\boldsymbol{B_n} - \eta \frac{\partial J_1}{\partial \boldsymbol{B_n}})$
                    
                \EndFor
                \State $\boldsymbol{\lambda} \gets \text{ProjectOntoSimplex}(\boldsymbol{\lambda} - \eta \frac{\partial J_1}{\partial \boldsymbol{\lambda}})$
                \State $J_{q} \gets J_1(.)$, $q \gets q + 1$
                \If {$|J_{q} - J_{q-1}| < \epsilon$ OR $q == \text{max iterations}$}
                    \State converged $\gets$ True
                \EndIf
            \EndWhile
            \State Return $\boldsymbol{\{\boldsymbol{B_n}\}},\boldsymbol{\lambda}$
	\end{algorithmic} 
\end{algorithm}

\subsection{Computation of $\boldsymbol{R_{j,k}}$ matrices}
For given $j,k\in [N]$ and $\boldsymbol{X}\in \mathbb{R}^{L_j\times L_k}$ we argued
$$\boldsymbol{R_{j,k}}\text{vec}(\boldsymbol{X}) = \mathcal{R}(\boldsymbol{\mathcal{D}_{j}X\mathcal{D}_k^T})$$
for a suitable $\boldsymbol{R_{j,k}}\in \mathbb{R}^{b\times L_jL_k}$, which we wish to compute. We will assume that we can generate samples from each column in $\boldsymbol{\mathcal{D}_n}\ \forall n\in [N]$. We define the outer product of functions $f_1:\mathbb{R}\rightarrow\mathbb{R}$ and $f_2:\mathbb{R}\rightarrow\mathbb{R}$ as $f_1\circ f_2 :\mathbb{R}^2\rightarrow\mathbb{R}^2$ with $(f_1 \circ f_2)(x_1,x_2) = f_1(x_1)f_2(x_2)\ \forall x_1, x_2 \in \mathbb{R}$. Recall that each column of our dictionaries is a 1D PDF and the operator $\mathcal{R}(.)$ takes a 2D density as its input. Thus we have 
$$\boldsymbol{\mathcal{D}_j X \mathcal{D}_k^T} = \sum_{l_j \in [L_j],l_k\in [L_k]}(\boldsymbol{\mathcal{D}_j}(:,l_j)\circ \boldsymbol{\mathcal{D}_k}(:,l_k))X(l_j,l_k)$$
Since $\mathcal{R}(.)$ is linear, we have:
$$\mathcal{R}(\boldsymbol{\mathcal{D}_{j}}\boldsymbol{X}\boldsymbol{\mathcal{D}_k}^T)=\sum_{l_j \in [L_j],l_k\in [L_k]} \mathcal{R}(\boldsymbol{\mathcal{D}_j}(:,l_j) \circ \boldsymbol{\mathcal{D}_k}(:,l_k))X(l_j,l_k)$$
$$=\sum_{l_j \in [L_j],l_k\in [L_k]} \boldsymbol{R_{j,k}}(:,(l_j,l_k))X(l_j,l_k).$$
In the above summation, each column of $\boldsymbol{R_{j,k}}$ is indexed with an ordered pair $(l_j,l_k),\ l_j \in [L_j],\ l_k\in [L_k]$ corresponding to the $(l_j,l_k)$th position in $\boldsymbol{X}$. Thus the $(l_j,l_k)$th column of $\boldsymbol{R_{j,k}}$ is given by $\boldsymbol{R_{j,k}}(:,(l_j,l_k))=\mathcal{R}(\boldsymbol{\mathcal{D}_j}(:,l_j) \circ \boldsymbol{\mathcal{D}_k}(:,l_k))$.\\

Given two 1D PDFs $f_1(.)$ and $f_2(.)$ from which we can generate samples, we can stochastically approximate $\mathcal{R}(f_1\circ f_2)$ by generating a large number of i.i.d. samples of $(X_1, X_2)$ with $X_1\sim f_1(.)$ and $X_2\sim f_2(.)$ where $X_1$ and $X_2$ are independent. This generates i.i.d samples of the 2D density $f_1 \circ f_2$. We can then project these samples along the given $M$ projection directions. We use the same binning as $\boldsymbol{y_{j,k}}$ and count the number of projected samples in each bin. We normalise this count by the total number of samples to obtain an approximation of the Radon transform of the 2D density. 

\begin{figure*}[h!]
\centering
\begin{subfigure}{0.45\textwidth}
    \includegraphics[width=\textwidth]{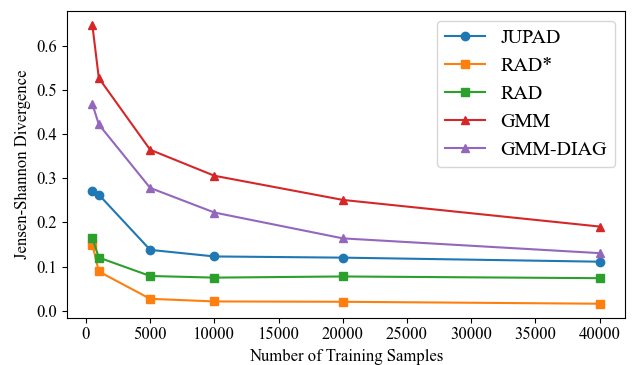}
    \caption{Gaussian dictionaries}
    \label{fig:first}
\end{subfigure}
\hfill
\begin{subfigure}{0.45\textwidth}
    \includegraphics[width=\textwidth]{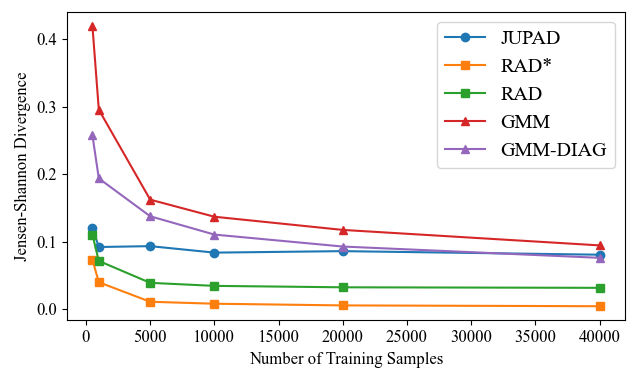}
    \caption{Laplacian dictionaries}
    \label{fig:second}
\end{subfigure}
\hfill
\begin{subfigure}{0.45\textwidth}
    \includegraphics[width=\textwidth]{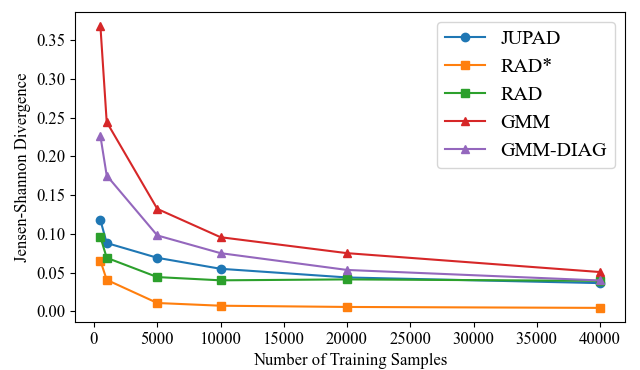}
    \caption{Mixture of Gaussian and Laplacian dictionaries}
    \label{fig:third}
\end{subfigure}
\hfill
\begin{subfigure}{0.45\textwidth}
    \includegraphics[width=\textwidth]{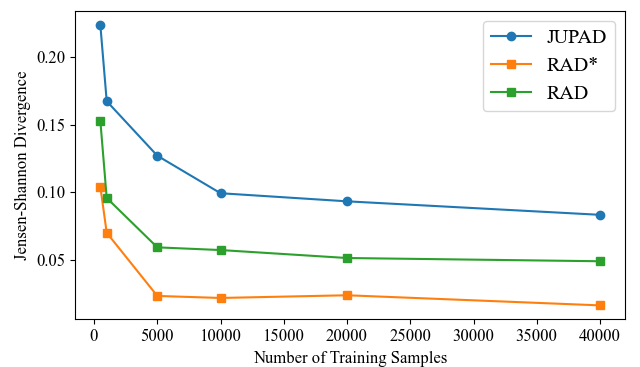}
    \caption{RV with continuous and discrete features}
    \label{fig:third}
\end{subfigure}
        
\caption{Jensen-Shannon Divergence versus Number of Training Samples $K$ for \textsc{Rad} (dictionaries constructed by observing 1D histograms), \textsc{Rad*}, \textsc{Jupad}, \textsc{Gmm} and \textsc{Gmm-Diag} (with diagonal covariance)}
\label{fig:figures}
\end{figure*}

\section{Numerical Results}
In this section we present several probability density estimation results on synthetic data and compare performance with previous algorithms. To generate artificial data, we first initialised a set of dictionaries with randomly chosen parameters and random mode factors to construct our CPD model. We sampled from this synthetic density by first choosing a random value $f \in [F]$ from the PMF given by $\boldsymbol{\lambda}$. Then for each feature $X_n$, we sampled from the dictionary column $\boldsymbol{\mathcal{D}_n}(:,l_n)$ where $l_n \in [L_n]$ was randomly sampled using the PMF given by $\boldsymbol{B_n}(:,f)$.

In order to evaluate algorithm performance, we computed the Jensen-Shannon Divergence (JSD) between the estimate and the true density. The JSD between two probability distributions $P$, $Q$ defined on $\mathbb{R}^d$ is defined as
$$JSD(P||Q) = \frac{1}{2}(D(P||M)+D(Q||M)),$$
where $M=\frac{1}{2}(P+Q)$ and $D(P||Q)$ is the Kullback-Leibler Divergence (KLD) between $P$ and $Q$. To compute the KL Divergence (KLD) we generate $S$ i.i.d. samples $\{x_i:i \in [S]\}$ of $X \sim P(.)$ with $S \gg 1$. We then approximate $D(P||Q)$ as:
$$D(P||Q) = \mathbb{E}_{X\sim P(.)}[\text{log}\frac{P(X)}{Q(X)}]\approx \frac{1}{S}\sum_{i \in [S]}\text{log}\frac{P(x_i)}{Q(x_i)}.$$
We compared the performance of our algorithm against \textsc{Jupad} \cite{shaan}, which uses dictionaries for continuous density estimation. The key advantage of our algorithm over \textsc{Jupad} was the use of 1D marginals instead of 2D, which greatly lowers sample complexity. We also compared performance with the Expectation Maximisation algorithm for Gaussian Mixture Models with full covariance matrices (referred to as \textsc{GMM}) and diagonal covariance matrices (referred to as \textsc{GMM-Diag}).

We do not present comparisons against the sinc-interpolation technique in \cite{sinc} because: (1) sinc-iterpolation to obtain the CDF does not guarantee a valid PDF as the resulting PDF may take negative values, (2) \textsc{JUPAD} \cite{shaan} is able to outperform it on synthetic density estimation when averaged absolute log likelihood ratio is used as the performance metric. 

We created multiple synthetic datasets and trained each algorithm on a random subset of $K$ data samples for 4 trials, and calculated the average JSD over these trials. We varied $K$ from 500 to 40000 samples.

We evaluated \textsc{Rad} under two settings: (1) \textsc{Rad*}: The true dictionaries used for generating the synthetic density were used by the algorithm, (2) \textsc{Rad}: The dictionaries were unknown and chosen by observing the histograms of 1D marginals of the generated  samples (as described in \cite{shaan}). Results for \textsc{Jupad} are only presented for the case where true dictionaries were used. Knowledge of the true dictionaries improved performance for both \textsc{Rad} and \textsc{Jupad}. While computing marginals via histogramming we used $K^{\frac{1}{3}}$ bins for \textsc{Rad}, where $K$ is the size of training data. For \textsc{Jupad}, we kept the number of bins fixed (50-100 depending on the dataset) with $K$ as it produced best results. We choose the number of components in both \textsc{Gmm} (full covariance) and \textsc{Gmm-Diag} (diagonal covariance) from 20 to 300 in steps of 20 by cross validation using negative log-likelihood. We used a $10\%$ subset of the training data for validation and fit the GMM on the remaining $90\%$.

We experimented with the following families of densities.
\subsubsection{Gaussian Dictionaries}
We have $N = 8$ features, and the tensor rank is $F = 25$. For each feature the dictionary $\boldsymbol{\mathcal{D}_n}$ contains $L_n = 10$ Gaussians, each with mean $\mu_{n,i} \sim \mathcal{U}(-1,1)$ and standard deviation $\sigma_{n,i} \sim \mathcal{U}(0.05, 0.2)$, chosen randomly. Here $\mathcal{U}(a,b)$ represents a uniform distribution with range $(a,b)$. In this and following experiments, all features share the same dictionary for simplicity, but they can also be chosen to be distinct.

\subsubsection{Laplacian Dictionaries}
Similar to the above experiment, we have $N = 6$ features, and the tensor rank is $F =7$. For each feature, the dictionary $\boldsymbol{\mathcal{D}_n}$ contains $L_n = 10$ Laplacians, each with mean $\mu_{n,i} \sim \mathcal{U}(-1,1)$ and standard deviation $\sigma_{n,i} \sim \mathcal{U}(0.05, 0.2)$.

\subsubsection{Mixture of Laplacians and Gaussians in each Dictionary}
In this experiment, we have $N = 5$ features, and the tensor rank is $F = 9$. For each feature, the dictionary $\boldsymbol{\mathcal{D}_n}$ contains $L_{n} = 12$ distributions of which 6 are Laplacians and 6 are Gaussians, each with mean $\mu_{n,i} \sim \mathcal{U}(-1,1)$ and standard deviation $\sigma_{n,i} \sim \mathcal{U}(0.05, 0.2)$.

\subsubsection{Mixture of Continuous and Discrete Features}
This experiment aims to capture the structure of most real-world datasets, which typically have mixed features. We have $N = 7$ features, and the tensor rank is $F = 13$. The first 4 features are continuous and are a mixture of $L_{n} = 8$ densities of which 4 are Laplacians and 4 are Gaussians, each with mean and standard deviation chosen as above. The last 3 features are discrete. Each one can take 4 distinct values in $S = \{ -1.5,-0.5,0.5,1.5\}$. The dictionary for a discrete feature consists of distributions of constant random variables, each assuming one of the constant value in $S$. We do not compare with GMMs for this case as they cannot represent discrete features.

In Fig.~\ref{fig:figures} we plot the JSD vs $K$ for each algorithm, for various density families. We see that \textsc{RAD*} achieves the lowest JSD in both low and high sample regimes. The performance of \textsc{RAD*} is expected to be better than \textsc{RAD}, as the former makes use of true dictionaries. However, even when we construct dictionaries by observing the 1D marginal histograms in \textsc{RAD}, it outperforms \textsc{JUPAD} which also uses true dictionaries. \textsc{RAD} also results in lower values of the JSD compared to \textsc{GMM} and \textsc{GMM-DIAG} for almost all sample complexity, across all the families of densities. Thus, our experiments demonstrate the superior sample complexity of joint density estimation for \textsc{RAD} over other algorithms.

\section*{Conclusion and Future Work}
We extend the work on learning a low-rank tensor representation of probability densities by combining the ideas of dictionaries and Radon transforms to develop a novel algorithm. Our model has minimal structural assumptions on the density and dictionaries enable us to represent various families of densities. We are able to achieve lower sample complexity than previous approaches using 1D marginals. Among structure-free methods our method performs well in the low sample regime as seen by experimental results. Future work may extend to adaptively learning dictionaries from data instead of selecting them through observation.

\end{document}